\documentclass[lettersize,journal]{IEEEtran}
\usepackage{amsmath,amsfonts}
\usepackage{algorithmic}
\usepackage{algorithm}
\usepackage{array}
\usepackage{textcomp}
\usepackage{stfloats}
\usepackage{url}
\usepackage{verbatim}
\usepackage{graphicx}
\usepackage{cite}

\usepackage{pdfpages}
\usepackage{mathrsfs}
\usepackage{amsfonts}
\usepackage{amsmath}
\usepackage{tikz}
\usepackage{amssymb}
\usepackage{textcomp}
\usepackage{comment}
\usepackage{multirow}
\usepackage{epstopdf}
\usepackage[switch]{lineno}
\usepackage{color}
\usepackage[colorlinks, linkcolor=red,  anchorcolor=blue, citecolor=black]{hyperref}

\usepackage{subfigure}

\begin{document}

\title{Few-Shot Learning  Meets Transformer: Unified Query-Support Transformers for Few-Shot Classification}

\author{Xixi Wang, Xiao Wang, \emph{Member, IEEE}, Bo Jiang*, Bin Luo, \emph{Senior Member, IEEE}
\thanks{The authors are all from School of Computer Science and Technology, Anhui University, Hefei 230601, China}
\thanks{Corresponding author: Bo Jiang}
}

\markboth{Journal of \LaTeX\ Class Files,~Vol.~14, No.~8, August~2021}%
{Shell \MakeLowercase{\textit{et al.}}: A Sample Article Using IEEEtran.cls for IEEE Journals}


\maketitle

\begin{abstract}
Few-shot classification which aims to recognize unseen classes using very limited samples has attracted more and more attention. Usually, it is formulated as a metric learning problem. The core issue of few-shot classification is how to learn (1) consistent representations for images in both support and query sets and (2) effective metric learning for images between support and query sets. In this paper, we show that the two challenges can be well modeled simultaneously via a unified \textbf{Q}uery-\textbf{S}upport Trans\textbf{Former} (\textbf{QSFormer}) model. To be specific, the proposed QSFormer involves \emph{global} query-support sample Transformer (sampleFormer) branch and \emph{local} patch Transformer (patchFormer) learning branch.
sampleFormer aims to capture the dependence of samples in support and query sets for image representation.
It adopts the Encoder, Decoder and Cross-Attention to respectively model the Support, Query (image) representation and Metric learning for few-shot classification task.
Also, as a complementary to global learning branch, we adopt a local patch Transformer to extract structural representation for each image sample by capturing the long-range dependence of local image patches. In addition, a novel Cross-scale Interactive Feature Extractor (CIFE) is proposed to extract and fuse multi-scale CNN features as an effective backbone module for the proposed few-shot learning method. All modules are integrated into a unified framework and trained in an end-to-end manner. Extensive experiments on four popular datasets demonstrate the effectiveness and superiority of the proposed QSFormer.
\end{abstract}


\section{Introduction}
\IEEEPARstart{C}{urrent} deep neural networks learn from large-scale training samples and achieve good performance on many tasks. However, in many scenarios, data collection and annotation is expensive and it is usually very challenging to collect enough data for the training of deep neural networks.
The Few-shot classification aims to recognize unseen/query classes by using very limited seen/support samples has attracted more and more attention.
%
\begin{figure}
\centering
\noindent\makebox[\textwidth][l] {
\includegraphics[width=3.3in]{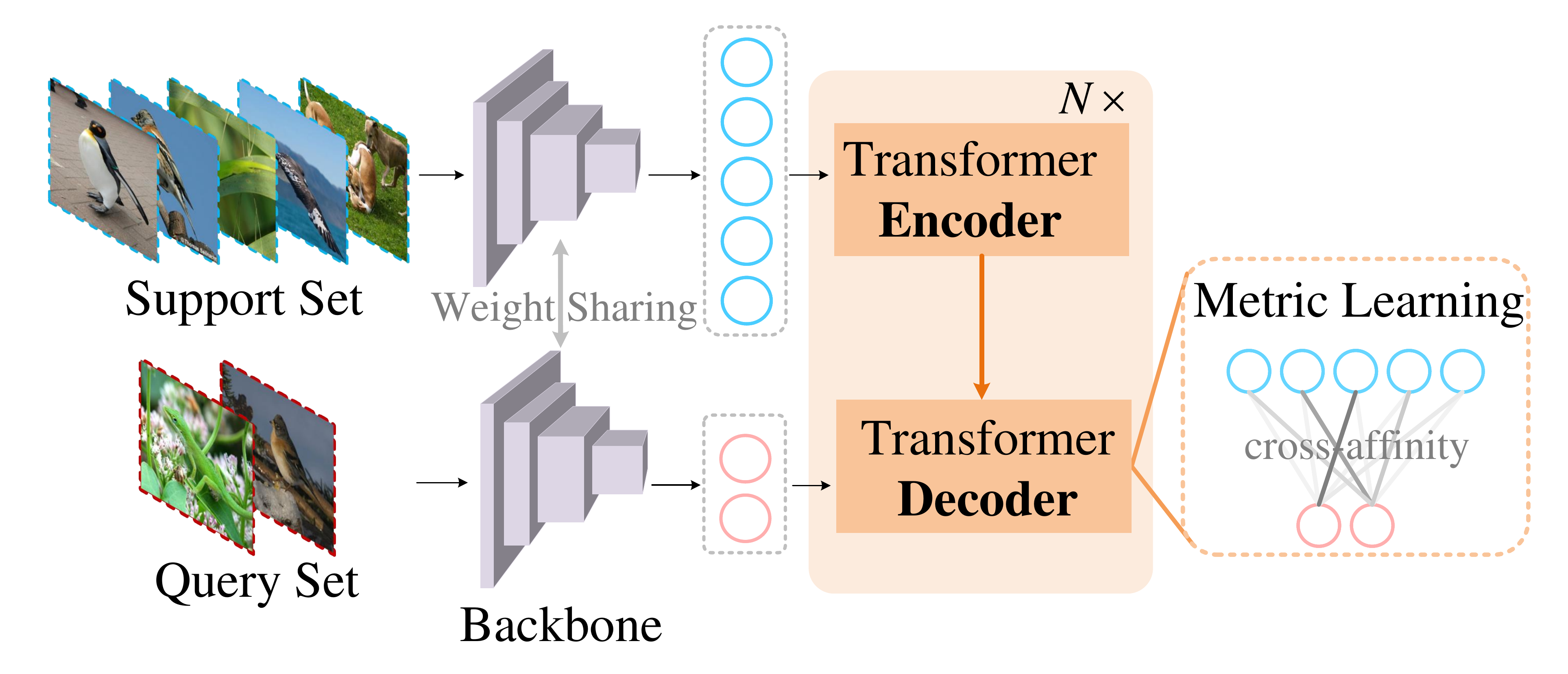}
}
\caption{ Illustration of our proposed unified Query-Support Transformer for few-shot learning.
It models the feature engineering on query/support samples and metric learning simultaneously. }
\label{fig:Fig.1}
\end{figure}

Many deep learning methods~\cite{rusu2018LEO,jamal2019task,zhang2020DeepEMD} have been proposed to address few-shot learning problem.
These methods can be roughly classified into three types, i.e., generation-based methods, optimization-based methods and metric-based methods.
Metric-based methods are derived to distinguish support and query samples by using some image representation and metric
learning techniques. 
As we know, the core issues for metric-based few-shot classification are two aspects: 1) How to learn consistent representations for images in both support and query sets. 2) How to conduct  effective metric learning for images between support and query sets.
According to our observation, existing works~\cite{vinyals2016MatchNet, snell2017ProtoNet, chen2019cosine, zhang2020DeepEMD, chen2021meta} usually first employ Convolution Neural Networks (CNNs) to learn image feature representation and then use a metric function to directly compute the similarities (e.g., cosine) between query and support images for few-shot classification.  
The good performance can be achieved, however, many recent studies~\cite{dosovitskiy2020vit,peng2021conformer} demonstrate that CNN only captures the local relations well due to its limited receptive field.
To address this issue, some researchers~\cite{he2022attribute, dong2022self, zhmoginov2022HT} propose to combine or replace CNN with  Transformer networks to model the long-range relationships of local image patches and obtain better image representation results.
{However, they may still obtain sub-optimal performance due to the following two reasons: }
1) Existing works generally adopt Transformers (or CNN+Transformer) as the backbone network for engineering each image representation, which obviously ignores the inherent relationships among samples in query and support sets for image representation.
2) Existing works generally adopt the two-stage learning scheme, i.e., `representation learning + metric learning'.
Although the two stages are usually learned together in an end-to-end manner, this decoupling way may lead to sub-optimal learning results.

%
%

To address these challenges,
in this work, we 
propose a unified Query-Support Transformer architecture for few-shot learning, termed QSFormer.
The core of QSFormer is our new design of query-support sample Transformer (named sampleFormer) module, which
aims to explore the relationships of samples for coupling \textbf{sample representations} and \textbf{metric learning of samples} together in a unified module for few-shot classification.
%
To be specific, as shown in Figure~\ref{fig:Fig.1}, we dexterously adopt the \emph{{Encoder}, {Decoder and {Cross-Attention} in our sampleFormer architecture to model the {Support}, {Query (image) representation} and {Metric learning} in few-shot classification task}}, respectively.
For the support branch, we represent all support images as a sequence of image tokens and feed them into the Transformer encoder to  enhance the support features.
For the query branch, it receives a sequence of query image tokens to learn their representations. Meanwhile, it interacts with the previous support branch via the cross-attention for modeling the similarities/affinities between query and support tokens, therefore, naturally achieving metric learning in the decoding procedure.

\begin{figure*}[!htb]
\centering
\includegraphics[width=1\textwidth]{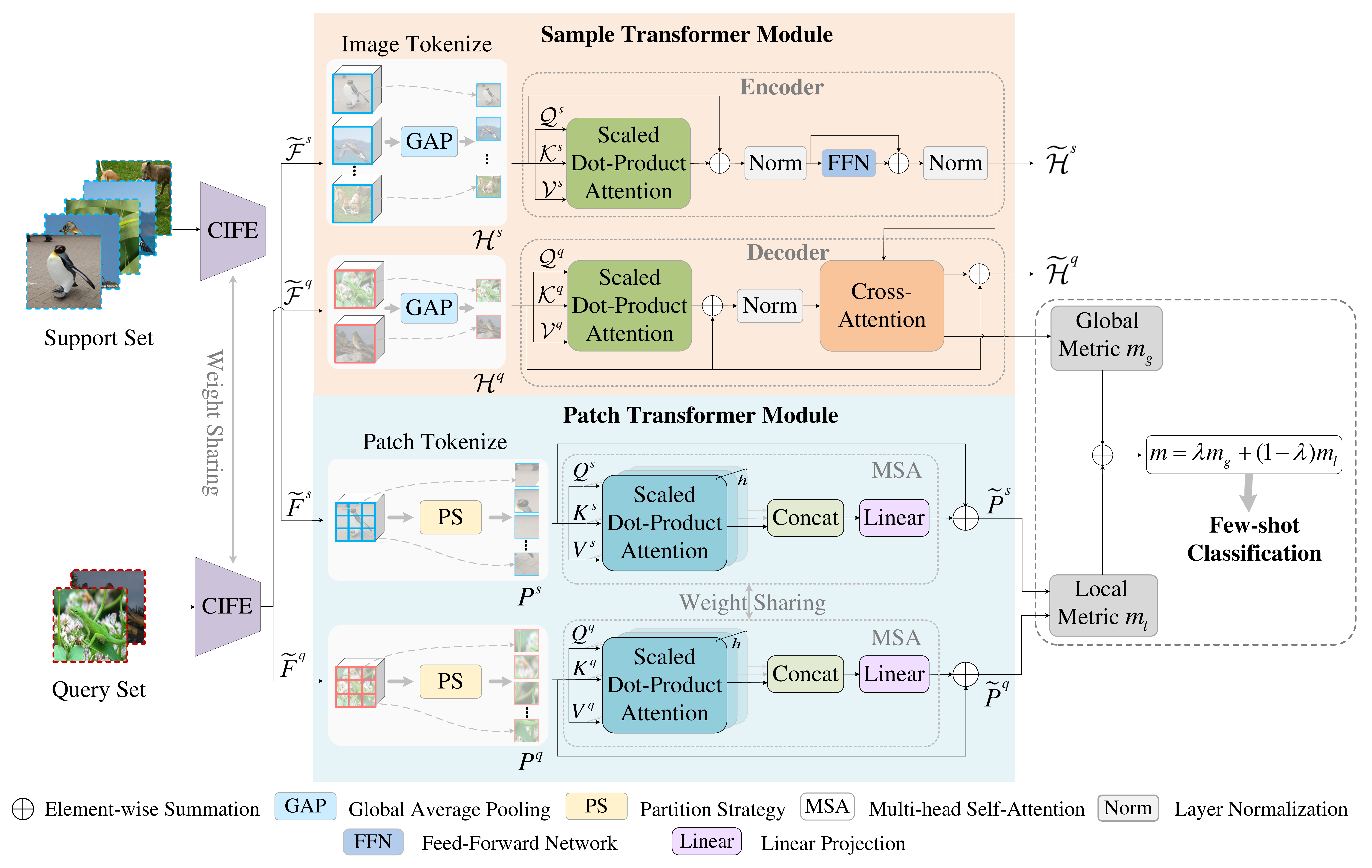}
\caption{An overview of the proposed QSFormer framework, which mainly consists of Cross-scale Interactive Feature Extractor (CIFE), Sample Transformer Module, Patch Transformer Module, Metric Learning and Few-shot Classification.
More details can be found in Section \uppercase\expandafter{\romannumeral3}.}
\label{fig:Fig.2}
\end{figure*}

Based on our newly proposed sampleFormer, we further extend it by introducing two additional new modules for high-performance few-shot learning, including Cross-scale Interactive Feature Extractor (CIFE) and local patch Transformer (patchFormer) module.
Specifically, as shown in Figure~\ref{fig:Fig.2}, given the query and support images, we first use CIFE as the backbone module to extract the image features. Then, the sampleFormer takes the embedded image tokens as input and outputs global metrics. Meanwhile, the local/patch correspondence of query-support image pairs is also considered using the patchFormer. The global and local metrics are combined for few-shot classification. Note that, the whole network can be optimized in an end-to-end way.

To sum up, the contributions of this paper can be summarized as follows:
\begin{itemize}
\item We propose a unified Query-Support Transformer (termed QSFormer) for few-shot learning, which models the representation learning and metric learning simultaneously.

\item We propose a novel Sample Transformer module (sampleFormer) to capture the sample relationships in few-shot problem setting. Also, we propose a patch Transformer (patchFormer) module for few-shot image representation and metric learning.
\item We propose a Cross-scale Interactive Feature Extractor  for image representation by considering the interaction of different CNN levels.

\item Extensive experiments on four widely used few-shot classification datasets demonstrate the effectiveness and superiority of our proposed method.

\end{itemize}

\section{Related Work}

\textbf{Few-shot Learning.}
Current few-shot learning algorithms can be broadly divided into two categories: optimization-based approaches~\cite{chen2019cosine,jamal2019task} and metric-based approaches~\cite{vinyals2016MatchNet,hou2019CAN,zhang2020DeepEMD,Xie2022DeepBDC}.
Our method is more relevant to the metric-based approaches, which mainly focus on the representation learning and metric learning of samples.
Specifically,
Sung et al.~\cite{sung2018RelationNet} propose a Relation Network (RN) for few-shot learning, which computes the relation scores between query examples and the few examples of each new class to classify the examples of new classes.
Hou et al.~\cite{hou2019CAN} develop a Cross Attention Network, which highlights the target object regions to enhance the feature representation by producing cross attention maps for each feature.
Zhang et al.~\cite{zhang2020DeepEMD} introduce Earth Mover's Distance to capture a structural distance between the local image representations for few-shot classification.
Xie et al.~\cite{Xie2022DeepBDC} introduce a deep Brownian Distance Covariance approach to learn image representations and then use distance metric for classification. \

\noindent
\textbf{Transformer for Few-shot Classification.}
Transformer ~\cite{vaswani2017attention} has universal modeling capability because its core module self-attention learning mechanism. 
%
%
In recent years, Transformer has been employed by a large number of researchers for various visual tasks, including object tracking~\cite{wang2021transformer, yu2022relationtrack}, object detection~\cite{carion2020end, guan2022m3detr}, object re-identification~\cite{liao2021TransMatcher, jia2022learning}, multi-label classification~\cite{liu2021query2label, chen2022sst}, Medical Image Segmentation~\cite{valanarasu2021medical, petit2021u}, and so on.
For few-shot learning tasks, some works~\cite{ye2019FEAT, Liu2021URT, jiang2022rgtransformer, he2022attribute, zhmoginov2022HT, dong2022self} demonstrate that Transformer architecture is also promising.
For example, Ye et al.~\cite{ye2019FEAT} develop a Few-Shot Embedding Adaptation Transformer (FEAT) to instantiate set-to-set transformation and thus make instance embedding task-specific for few-shot learning.
Liu et al.~\cite{Liu2021URT} propose a Universal Representation Transformer (URT) layer by combining feature representations from multiple domains together for multi-domain few-shot classification.
%
%
%
Zhmoginov et al.~\cite{zhmoginov2022HT} introduce a transformer-based model, called HyperTransformer (HT), which encodes task-dependent variations in the weights of a small CNN model for few-shot learning.
These works mainly employ Transformer architecture for representation learning.
Differently, in our work, we develop a Query-Support Transformer (QSFormer) to accomplish both feature representation and metric learning simultaneously.

\section{The Proposed Method}
\label{sec:sec3}
The purpose of few-shot classification is to classify the unseen samples when only a small number of samples are available.
Many recent approaches~\cite{hou2019CAN,zhang2020DeepEMD,fei2021MELR,Wang2022TSNPM} indicate that the episode mechanism provides an effective way for few-shot classification task and we follow them in both training  and testing phases.
Formally, let $\mathcal{D}_{train}$, $\mathcal{D}_{val}$ and $\mathcal{D}_{test}$ respectively represent meta-training, meta-validation and meta-testing set, where $\mathcal{D}_{train} \cap \mathcal{D}_{val} \cap \mathcal{D}_{test} = \emptyset$.
Taking $C$-way $K$-shot few-shot classification task as an example, each episode consists of support set $\mathcal{X}^s=\{(X^s_i, Y^s_i) \}^{n_s}_{i=1}$ and query set $\mathcal{X}^q=\{(X^q_j, Y^q_j) \}^{n_q}_{j=1}$.
Concretely, we randomly select $C$ classes and $K$ labeled samples per class to form the support set $\mathcal{X}^s$, i.e., $n_s=C \times K$.
Meanwhile, we randomly sample $q$ samples per class to form the query set $\mathcal{X}^q$, i.e., $n_q=C \times q$.
%

As shown in Figure~\ref{fig:Fig.2}, we propose a novel Query-Support Transformer (QSFormer) framework for few-shot learning, which contains the following four parts:
%
\begin{itemize}
 \item  \textbf{Cross-Scale Interactive Feature Extractor (CIFE):} we propose a cross-scale interactive feature extractor as backbone network to obtain the spatial enhanced support/query CNN feature representations.

 \item \textbf{Sample Transformer Module:} we introduce a query-support sample Transformer (sampleFormer) module to couple image sample representation and global metric learning of samples together for few-shot learning.

\item \textbf{Patch Transformer Module:} we also propose a patch Transformer (patchFormer) module to model the context correlation of patches in each image sample to conduct the local metric learning between query-support sample pairs.

 \item \textbf{Metric Learning and Few-shot Classification:} we acquire the final metric by combining global metric obtained via sampleFormer and local metric obtained via patchFormer together and final achieve few-shot classification.

\end{itemize}
Below, we introduce the details of these modules.

\subsection{Cross-scale Interactive Feature Extractor}
\label{subsec:ss1}
We introduce a novel Cross-scale Interactive Feature Extractor (CIFE) as backbone module, which aims to obtain the ego-context CNN feature representations for support and query samples.

\begin{figure}[!t]
\centering
\includegraphics[width=0.49\textwidth]{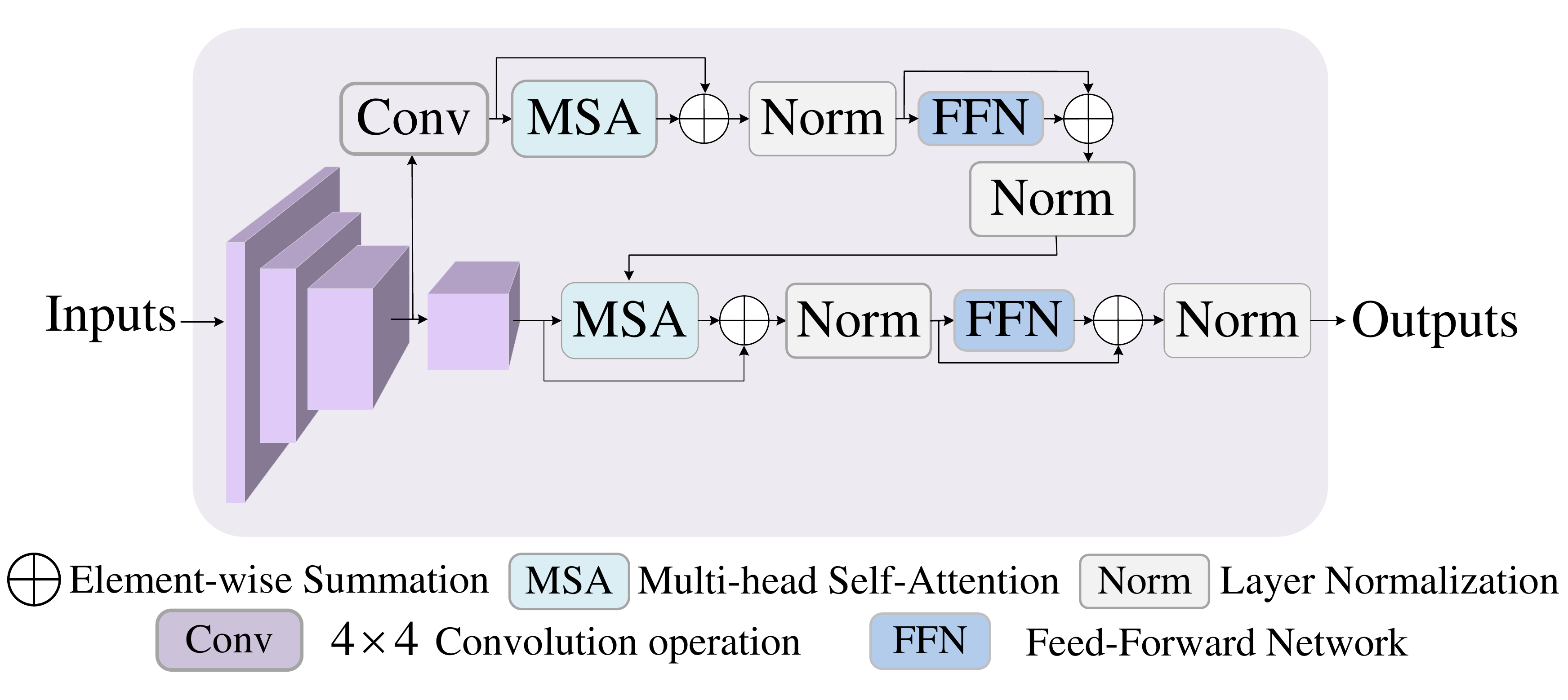}
\caption{Illustration of Cross-scale Interactive Feature Extractor (CIFE) for feature extraction.
}
\label{fig:Fig.3}
\end{figure}

As shown in Figure~\ref{fig:Fig.3}, taking the support image set $\mathcal{X}^s = \{X^s_1, X^s_2, ..., X^s_{n_s} \}$ as inputs, we first use the pre-trained ResNet-12 to generate the initial multi-scale feature representations $\mathcal{F}^s_l \in \mathbb{R}^{n_s \times c_l \times h_l \times w_l}, l \in \{1,2,3,4 \}$, where $n_s$ represents the number of support samples in each episode and $c_l$, $h_l$ and $w_l$ denote the channel, height and width of support feature map in the $l$-th level respectively.
Then, we employ a Transformer architecture~\cite{vaswani2017attention} consisting of multi-head self-attention (MSA), layer normalization (LN), feed-forward network (FFN) and residual connection to achieve the interaction of multi-scale features.
Finally, we can obtain the spatial enhanced feature representations for support samples as $\widetilde{\mathcal{F}}^s = \{\widetilde{F}^s_1, \widetilde{F}^s_2, \cdots, \widetilde{F}^s_{n_s} \} \in \mathbb{R}^{n_s \times c \times h \times w}$.
Similarly,  we obtain the spatial enhanced features for query samples as $\widetilde{\mathcal{F}}^q = \{\widetilde{F}^q_1, \widetilde{F}^q_2, \cdots, \widetilde{F}^q_{n_q} \} \in \mathbb{R}^{n_q \times c \times h \times w}$.
The parameters of CIFE are shared for support and query branches.
In practice, we empirically set $c=640$ and $h=w=5$.
%

\subsection{Sample Transformer Module}
\label{subsec:ss2}
To achieve both image sample representation and metric learning of samples in a unified module, we design a novel query-support sample Transformer module, named sampleFormer.
The proposed sampleFormer mainly consists of Encoder and Decoder, as shown in Figure~\ref{fig:Fig.2}.

\textbf{Encoder.}
The purpose of the Encoder is to mine the relationships of samples in support set to obtain better support feature representations.
To this end, based on the aforementioned support features $\widetilde{\mathcal{F}}^s \in \mathbb{R}^{n_s \times c \times h \times w}$, we first introduce \emph{image tokenize}, which utilizes a global average pooling 
and reshape operation to gain the token sequence $\mathcal{H}^s = \{H^s_1, H^s_2, \cdots, H^s_{n_s} \}\in \mathbb{R}^{n_s \times c}$ of support samples, where each token $H^s_i$ denotes a support image sample.
As shown in Figure~\ref{fig:Fig.2}, we can see that the main component of encoder is \emph{attention mechanism}, whose inputs are Query $\mathcal{Q}^s \in \mathbb{R}^{n_s \times c}$, Key $\mathcal{K}^s \in \mathbb{R}^{n_s \times c}$, and Value $\mathcal{V}^s \in \mathbb{R}^{n_s \times c}$ obtained by conducting three linear projections on $\mathcal{H}^s$ respectively.
Next, it employs dot-product operation to obtain a correlation/affinity matrix $Attn_{s \to s}(\mathcal{Q}^s,\mathcal{K}^s)$ of different support samples as 
\begin{equation}
Attn_{s \to s}(\mathcal{Q}^s,\mathcal{K}^s) = Softmax(\frac{\mathcal{Q}^{s}(\mathcal{K}^{s})^T}{\sqrt{c}})
\label{equ:eq1}
\end{equation}
where $c$ denotes the dimension of support features.
It learns the representations for support samples by conducting the message passing operation as 
\begin{equation}
\widehat{\mathcal{H}}^s = LN(\mathcal{H}^s + Attn_{s \to s}(\mathcal{Q}^s,\mathcal{K}^s)\mathcal{V}^s)
\label{equ:eq2}
\end{equation}
where $LN(\cdot)$ refers to layer normalization.
Besides, we add Feed-Forward Network (FFN)~\cite{dosovitskiy2020vit} and residual operation to obtain the final support sample representations as,
\begin{equation}
\widetilde{\mathcal{H}}^s = LN(\widehat{\mathcal{H}}^s + FFN(\widehat{\mathcal{H}}^s))
\label{equ:eq3}
\end{equation}
where $\widetilde{\mathcal{H}}^s = \{\widetilde{H}^s_1, \widetilde{H}^s_2 \cdots, \widetilde{H}^s_{n_s} \} \in \mathbb{R}^{n_s \times c}$.
$n_s$ denotes the number of support samples and $c$ is the feature dimension.
FFN consists of two fully-connection layers. 

\textbf{Decoder.}
The Decoder aims to explore the dependence of samples in query set to learn the representations for query samples and also mines the intrinsic metrics of samples in query and support sets.
To be specific,
it takes the aforementioned encoded support features $\widetilde{\mathcal{H}}^s \in \mathbb{R}^{n_s \times c}$ and query feature embeddings $\widetilde{\mathcal{F}}^q \in \mathbb{R}^{n_q \times c \times h \times w}$ as its inputs.
The \emph{image tokenize} is applied on $\widetilde{\mathcal{F}}^q$ to obtain the initial query token sequence $\mathcal{H}^q = \{H^q_1, H^q_2, \cdots, H^q_{n_q} \} \in \mathbb{R}^{n_q \times c}$, where each token $H^q_j$ denotes a query image sample.
Similar to the Encoder branch, we first leverage {self-attention message passing mechanism} to model the relationships among query samples and learn representations for query samples as
\begin{align}
&Attn_{q \to q}(\mathcal{Q}^q,\mathcal{K}^q) = Softmax(\frac{\mathcal{Q}^{q}(\mathcal{K}^{q})^T}{\sqrt{c}})\\
&\widehat{\mathcal{H}}^q = LN(\mathcal{H}^q + Attn_{q \to q}(\mathcal{Q}^q,\mathcal{K}^q)\mathcal{V}^q)
\label{equ:eq5}
\end{align}
where $LN(\cdot)$ denotes layer normalization.

Afterward, based on the support features $\widetilde{\mathcal{H}}^s$ and query features $\widehat{\mathcal{H}}^q$, we employ a \textbf{cross-attention} mechanism to explore the relationships between support and query samples for query sample representations.
Specifically, it first computes the cross-affinities between support and query samples as follows
%
\begin{align}
Attn_{q \to s}(\mathcal{Q}^q, \mathcal{K}^s) = Softmax(\mathcal{Q}^q(\mathcal{K}^s)^T)
\label{equ:eq6}
\end{align}
Then, it learns query sample representations by aggregating the information from support samples as follows
%
\begin{align}
\widetilde{\mathcal{H}}^q = \widehat{\mathcal{H}}^q + LN(Attn_{q \to s}(\mathcal{Q}^q, \mathcal{K}^s)\mathcal{V}^s)
\label{equ:eq7}
\end{align}
where $\widetilde{\mathcal{H}}^q \in \mathbb{R}^{n_q \times c}$ and
$LN(\cdot)$ denotes layer normalization.
$\mathcal{Q}^q \in \mathbb{R}^{n_q \times c}$ is computed by conducting a linear projection on $\widehat{\mathcal{H}}^q$.  $\mathcal{K}^s \in \mathbb{R}^{n_s \times c}$ and $\mathcal{V}^s \in \mathbb{R}^{n_s \times c}$ are obtained by conducting two different linear projections on $\widetilde{\mathcal{H}}^s$, respectively.

\textbf{Remark.}
The above cross-affinities $ Attn_{q \to s}(\mathcal{Q}^q, \mathcal{K}^s)$ naturally reflect the similarities/affinities between  support and query samples.
In our work, we regard them as global metric $m_g$ for all support and query samples, i.e.,
\begin{align}
m_g(\mathcal{X}^s,\mathcal{X}^q )=Attn_{q \to s}(\mathcal{Q}^q,\mathcal{K}^s)
\label{equ:eq8}
\end{align}
where $m_g(\mathcal{X}^s,\mathcal{X}^q)$ contains the similarities for all query-support sample pairs in each episode.
For convenience, in the following, we also use $m_g({X}^s,{X}^q)$ to denote the metric between image ${X}^s$ and ${X}^q$, where $X^s\in \mathcal{X}^s, X^q\in \mathcal{X}^q$.
We can utilize $m_g({X}^s,{X}^q)$ for query sample classification, as discussed in the following Section Metric Learning and Few-shot Classification.
Therefore,
we can note that both query/support \textbf{sample representation} and \textbf{metric learning} in few-shot learning task are conducted simultaneously in our sampleFormer architecture. This is one main aspect of the proposed sampleFormer module.



\subsection{Patch Transformer Module}

As a complementary to the above sampleFormer branch, we also develop a  query-support Patch Transformer Module (patchFormer) to capture the more visual content of each image sample for local metric. 
As shown in Figure~\ref{fig:Fig.2}, patchFormer mainly consists of multi-head self-attention (MSA) and residual connection. 
Here, we omit Feed-Forward Network used in regular Transformer~\cite{dosovitskiy2020vit} for simplicity consideration.  The parameters of MSA are shared on both support and query branches. 

%
%

Concretely, for each input support sample $X^s$ and query sample $X^q$, we first obtain their feature embedding $\widetilde{F}^s \in \mathbb{R}^{c \times h \times w}$ and $\widetilde{F}^q \in \mathbb{R}^{c \times h \times w}$ by using the above CIFE, followed by the \emph{patch tokenize}~\cite{dosovitskiy2020vit}
to obtain the initial patch token sequence for each support and query image, i.e., $P^s=\{p^s_1, p^s_2, \cdots, p^s_{hw}\} \in \mathbb{R}^{hw \times c}$ and $P^q = \{p^q_1, p^q_2, \cdots, p^q_{hw}\} \in \mathbb{R}^{hw \times c}$.
Then, we employ multi-head self-attention (MSA)~\cite{vaswani2017attention} with shared weights and residual operation to transform the support and query image patch features as
%
\begin{equation}
\begin{aligned}
\widetilde{P}^s &= LN(P^s + MSA(P^s)) \\
\widetilde{P}^q &= LN(P^q + MSA(P^q))
\label{equ:eq10}
\end{aligned}
\end{equation}
where $LN(\cdot)$ denotes layer normalization.

Based on the above patch representations $\widetilde{P}^s=\{\widetilde{p}^s_1,  \widetilde{p}^s_2, \cdots, \widetilde{p}^s_{hw}\}$ and $\widetilde{P}^q =\{\widetilde{p}^q_1,\widetilde{p}^q_2, \cdots, \widetilde{p}^q_{hw}\}$, we  then adopt the Earth Mover's
Distance (EMD)~\cite{hitchcock1941distribution,zhang2020DeepEMD} to compute their structural similarity.
It first computes the distance between all patch pairs $(\widetilde{p}^s_i, \widetilde{p}^q_j)$ and then acquires the optimal matching between patches of two images that have the minimum distance cost.
Finally, it returns the image-level metric by aggregating the metrics of all matched patch pairs.
In this paper, we denote this metric as local metric between support sample $X^s$ and query sample $X^q$, i.e.,
%
\begin{equation}
\begin{aligned}
& m_l(X^s, X^q) =  EMD(\widetilde{P}^q, \widetilde{P}^s)
\label{equ:eq11}
\end{aligned}
\end{equation}


\subsection{Metric Learning and Few-Shot Classification}
Given the support samples $(X^s, Y^s) \in \mathcal{X}^s$ with known labels and input query sample $X^q \in \mathcal{X}^q$, few-shot classification aims to determine the label of the query sample.
To achieve this task,
we first obtain the sample-based global metric $m_g(X^s, X^q)$ via Equ. (\ref{equ:eq8}) and patch-based local metric $m_l(X^s, X^q)$ via Equ. (\ref{equ:eq11}) respectively and combine them together to obtain the final metric/similarity between $X^s$ and $X^q$ as
\begin{equation}
m(X^s, X^q) = \lambda m_g(X^s, X^q) + (1-\lambda)m_l(X^s, X^q)
\label{equ:eq12}
\end{equation}
where $\lambda \in (0,1)$ is a tradeoff parameter.

Then, we can conduct few-shot classification by using the nearest neighbor classification strategy, i.e., the label of query $X^{q}$ is determined by the label $Y^{s^*}$ of the support sample $X^{s^*}$ that is most similar with query $X^{q}$, as used in previous works~\cite{vinyals2016MatchNet,zhang2020DeepEMD}.
%
%

\textbf{Loss Function.}
In the training phase, we employ two loss functions for the proposed QSFormer.
First, for the sampleFormer module, we specifically introduce a contrastive loss as suggested in work~\cite{oord2018infoNCE,liu2021learning}, which encourages the positive query-support sample pairs with same label (i.e., $Y^s = Y^q$) to be closing and the negative query-support sample pairs with different labels (i.e., $Y^s \neq Y^q$) are far away in each episode.
This loss function can be written as follows,
\begin{equation}
L_{cl}  = - log \frac{\sum\limits_{Y^s = Y^q} e^{m_g(X^s,X^q)} }{\sum\limits_{Y^s = Y^q} e^{m_g(X^s,X^q)}+\sum\limits_{Y^s \neq Y^q}e^{m_g(X^s,X^q)}}
\label{equ:eq14}
\end{equation}
where $m_g(X^s,X^q)$ is the global metric between query $X^q$ and support sample $X^s$.
The whole network is trained in an end-to-end way by minimizing the Cross-Entropy (CE) loss function $\mathcal{L}_{ce}$ ~\cite{zhang2020DeepEMD}.
Thus, the total loss function can be formulated as
\begin{equation}
\mathcal{L}_{total} = \alpha \mathcal{L}_{ce}(\hat{Y}^q, Y^q) + (1-\alpha) \mathcal{L}_{cl}
\label{equ:eq15}
\end{equation}
where $\hat{Y}^q$ is the label prediction obtained by our method and $Y^q$ denotes the corresponding ground-truth label.
$\alpha \in (0,1)$ is the balanced hyper-parameter.

\begin{table*}[!htb]
\centering
\caption{5-way result comparison of ours and state-of-the-art methods on miniImageNet and tieredImagenet datasets.
 Most results are from~\cite{zhang2020DeepEMD} or the original papers. The 1$^{st}$, 2$^{rd}$ and 3$^{rd}$ are respectively in \textcolor{red}{Red}, \textcolor{blue}{Blue} and \textcolor{green}{Green}. * denotes this method is reproduced with our settings.}
\begin{tabular}{l|l|ll|ll}
\hline
\multicolumn{1}{c|}{\multirow{2}{*}{Method}} &\multirow{2}{*}{Backbone} &\multicolumn{2}{c|}{miniImagenet}                   &\multicolumn{2}{c}{tieredImagenet} \\ \cline{3-6}
\multicolumn{1}{c|}{} &       &\multicolumn{1}{c}{1-shot}  &\multicolumn{1}{c|}{5-shot} &\multicolumn{1}{c}{1-shot}                        &\multicolumn{1}{c}{5-shot}  \\ \hline
DHL~\cite{Zhang2022DHL}          &Conv4     &61.99 $\pm\ -$    & 78.71 $\pm\ -$   &57.89 $\pm\ -$    &73.62 $\pm\ -$ \\
cosine classifier~\cite{chen2019cosine}* &ResNet12 &59.64 $\pm$ 0.27 &75.80 $\pm$ 0.21 &55.87 $\pm$ 0.31 &80.92 $\pm$ 0.23 \\  %
TADAM~\cite{oreshkin2018TADAM}   &ResNet12  &58.50 $\pm$ 0.30  &76.70 $\pm$ 0.30 &\multicolumn{1}{c}{$\ \ -$} &\multicolumn{1}{c}{$\ \ -$} \\
ECM~\cite{ravichandran2019ECM}   &ResNet12  &59.00 $\pm\ -$    &77.46 $\pm\ -$    &63.99 $\pm\ -$    &81.97 $\pm\ -$ \\  %
TPN~\cite{liu2019TPN}            &ResNet12  &59.46 $\pm\ -$    &75.65 $\pm\ -$    &59.91 $\pm$ 0.94  &73.30 $\pm$ 0.75 \\  %
ProtoNet~\cite{snell2017ProtoNet}*  &ResNet12  &63.03 $\pm$ 0.29  &78.72 $\pm$ 0.21  & 68.68 $\pm$ 0.34  &\textcolor{green}{85.09 $\pm$ 0.23} \\  %
MTL~\cite{sun2019MTL}            &ResNet12  &61.20 $\pm$ 1.80  &75.50 $\pm$ 0.80 &\multicolumn{1}{c}{$\ \ -$} &\multicolumn{1}{c}{$\ \ -$} \\  %
DC~\cite{lifchitz2019DC}         &ResNet12  &62.53 $\pm$ 0.19  &\textcolor{blue}{79.77 $\pm$ 0.19} &\multicolumn{1}{c}{$\ \ -$} &\multicolumn{1}{c}{$\ \ -$} \\  %
MetaOptNet~\cite{lee2019MetaOptNet} &ResNet12&62.64 $\pm$ 0.82  &78.63 $\pm$ 0.46  &65.99 $\pm$ 0.72  &81.56 $\pm$ 0.53\\ %
MatchNet~\cite{vinyals2016MatchNet}*  &ResNet12&61.24 $\pm$ 0.29  &73.93 $\pm$ 0.23  &\textcolor{green}{71.01 $\pm$ 0.33} &83.12 $\pm$ 0.24 \\ %
Meta-Baseline~\cite{chen2021meta} &ResNet12  &63.17 $\pm$ 0.23  &79.26 $\pm$ 0.17  &68.62 $\pm$ 0.27  &83.74 $\pm$ 0.18  \\ %
CAN~\cite{hou2019CAN}             &ResNet12  &63.85 $\pm$ 0.48  &\textcolor{green}{79.44 $\pm$ 0.34}  &69.89 $\pm$ 0.51 &84.23 $\pm$ 0.37 \\ %
PPA~\cite{qiao2018PPA}            &WRN-28-10 &59.60 $\pm$ 0.41  &73.74 $\pm$ 0.19  &65.65 $\pm$ 0.92  &83.40 $\pm$ 0.65 \\ %
wDAE-GNN~\cite{gidaris2019GNN}    &WRN-28-10 &61.07 $\pm$ 0.15  &76.75 $\pm$ 0.11  &68.18 $\pm$ 0.16  &83.09 $\pm$ 0.12 \\ %
LEO~\cite{rusu2018LEO}            &WRN-28-10 &61.76 $\pm$ 0.08  &77.59 $\pm$ 0.12  &66.33 $\pm$ 0.05  &81.44 $\pm$ 0.09 \\ %
FEAT~\cite{ye2019FEAT}*           &ResNet12  &\textcolor{green}{64.75 $\pm$ 0.28}  &79.96 $\pm$ 0.20  &\textcolor{blue}{71.34 $\pm$ 0.33}  &\textcolor{blue}{85.28 $\pm$ 0.23}  \\ %
HT~\cite{zhmoginov2022HT}         &Transformer &54.10 $\pm\ -$   &68.50 $\pm\ -$    &56.10 $\pm\ -$      &73.30 $\pm\ -$  \\ %
DeepEMD~\cite{zhang2020DeepEMD}*  &ResNet12  &\textcolor{red}{65.43 $\pm$ 0.28}  &79.28 $\pm$ 0.20 &69.84 $\pm$ 0.32 &84.06 $\pm$ 0.23  \\ %
DeepBDC~\cite{Xie2022DeepBDC}*    &ResNet12  &60.76  $\pm$ 0.28  &78.25 $\pm$ 0.20                   &63.03 $\pm$ 0.31
&81.57 $\pm$ 0.22 \\\hline %
QSFormer (Ours)   &ResNet12  &\textcolor{blue}{65.24 $\pm$ 0.28}  &\textcolor{red}{79.96 $\pm$ 0.20} &\textcolor{red}{72.47 $\pm$ 0.31}  &\textcolor{red}{85.43 $\pm$ 0.22}        \\ \hline
\end{tabular}
\label{table:t1}
\end{table*}

\textbf{Implementation Details.}
To achieve a fair comparison, the ResNet-12~\cite{chen2019cosine,zhang2020DeepEMD} with fully connected layers removed is adopted as the backbone module.
It is firstly pre-trained from scratch and then use the episodic training based on meta-learning framework by following works~\cite{chen2021meta,zhang2020DeepEMD}.
We empirically conduct the feature interaction of the last two levels in CIFE to obtain the enhanced sample features.
We randomly sample 50/1000/5000 episodes from the training/validation/testing set on four public datasets.
We compute the average accuracy and the corresponding 95$\%$ confidence interval to obtain the final performances of four datasets. Our proposed method is implemented by using Python  on a server with a single 11G NVIDIA 2080Ti GPU.
More hyper-parameter settings on four benchmarks for the proposed QSFormer are shown in Table~\ref{table:t7}.

\section{Experiments}

\subsection{Datasets and Evaluation Metric}
To verify our proposed QSFormer, we conduct extensive experiments on four publicly popular datasets for few-shot classification task, including \textbf{miniImageNet}~\cite{vinyals2016MatchNet}, \textbf{tieredImageNet}~\cite{ren2018tieredImageNet}, \textbf{Fewshot-CIFAR100}~\cite{oreshkin2018TADAM} and \textbf{Caltech-UCSD Birds-200-2011}~\cite{wah2011CUB}.
We also conduct cross-domain experiments to evaluate the domain transfer ability of the proposed model.
The recognition accuracy is adopted as the evaluation metric for our experiments.
More details of datasets description are as follow.

 \textbf{miniImageNet.} This dataset is a sub-dataset of ImageNet~\cite{russakovsky2015imagenet}.
 It contains a total of 100 classes with 600 samples in each class.
 As suggested in work~\cite{ravi2016optimization}, we divide these classes into training set, validation set and testing set, which respectively contains 64, 16 and 20 classes.

 \textbf{tieredImageNet.}
 It contains 608 classes from 34 super-classes, with a total of 779,165 samples. Following~\cite{ren2018tieredImageNet}, we split 34 super-classes into 20 super-classes (351 classes) for meta-training, 6 super-classes (97 classes) for meta-validation and 8 super-classes (160 classes) for meta-testing.

 \textbf{FC100.}
 Fewshot-CIFAR100 is built upon the CIFAR100 dataset for few-shot classification task.
 It's named FC100 for short hereafter.
 It contains a total of 60,000 images from 100 classes.
 To reduce the information overlap, we group the 100 classes into 20 super-classes by following work~\cite{oreshkin2018TADAM}.
 Then, we divide these super-classes into training set, validation set and testing set, which contains 12, 4 and 4 super-classes respectively.

 \textbf{CUB.} Caltech-UCSD Birds-200-2011 dataset is an extended vision of CUB-200 dataset.
 It's termed CUB for short hereafter.
 CUB is originally presented in fine-grained bird classification task.
 It contains the total of 11,788 images from 200 classes.
 As suggested by~\cite{ye2019FEAT}, we divide 200 classes into 100 classes for meta-training, 50 classes for meta-validation and 50 classes for meta-testing.

 \textbf{miniImageNet $\to$ CUB.}
By following~\cite{chen2019cosine}, we train a model on miniImageNet dataset and evaluate on the CUB dataset to verify the transfer ability of model.
In this experimental setting, specifically, we use all 100 classes of miniImageNet, with 600 samples per class for meta-training and use the meta-testing set (50 classes) of CUB dataset for meta-testing.

\subsection{Comparison with State-of-the-art Methods}
As shown in Table~\ref{table:t1}, we report our results and compare with other state-of-the-art (SOTA) approaches on miniImageNet~\cite{vinyals2016MatchNet} and tieredImageNet~\cite{ren2018tieredImageNet} datasets.
From this Table, we can find that the proposed QSFormer beats many SOTA models on the miniImageNet dataset.
For example, QSFormer exceeds the transformer-based HT~\cite{zhmoginov2022HT} method by +11.14\% and +11.46\% in 1-shot and 5-shot tasks, respectively.
For the attention mechanism based CAN~\cite{hou2019CAN}, our model also outperforms it on the 1-shot/5-shot task by +1.39$\%$/+0.52$\%$.
Compared with FETA~\cite{ye2019FEAT} that is also developed based on ResNet12 and Transformer, the proposed QSFormer has better results.

From Table~\ref{table:t1}, we can see that QSFormer achieves the best performance on the tieredImageNet dataset, i.e., 72.47$\pm$0.31 and 85.43$\pm$0.22 in 1-shot and 5-shot tasks. It exceeds the CAN~\cite{hou2019CAN} by +2.58 and +1.2 points in 1-shot and 5-shot tasks. Similar conclusions can also be drawn from the experimental results of Fewshot-CIFAR100~\cite{oreshkin2018TADAM} and CUB~\cite{wah2011CUB} datasets,
as illustrated in Table~\ref{table:t2} and Table~\ref{table:t3}.
All in all, the proposed QSFormer attains SOTA performance on multiple FSL datasets, which fully demonstrates the effectiveness and advantages of our proposed QSFormer model.

\begin{table}[!t]
\centering
\caption{5-way result comparison of ours and state-of-the-art methods on Fewshot-CIFAR100 dataset. The 1$^{st}$, 2$^{rd}$ and 3$^{rd}$ are respectively in \textcolor{red}{Red}, \textcolor{blue}{Blue} and \textcolor{green}{Green}. * denotes this method is reproduced with our settings.}
\begin{tabular}{l|ll}
\hline
Method  & $\ \ \ \ \ $1-shot       & $\ \ \ \ \ $5-shot       \\ \hline
cosine classifier~\cite{chen2019cosine}*   & 39.47 $\pm$ 0.23         & 56.29 $\pm$ 0.25 \\ %
FEAT~\cite{ye2019FEAT}*                   & 42.28 $\pm$ 0.26         & 56.37 $\pm$ 0.25 \\
TADAM~\cite{oreshkin2018TADAM}   & 40.10 $\pm$ 0.40         & 56.10 $\pm$ 0.40 \\ %
ProtoNet~\cite{snell2017ProtoNet}*  & 40.91 $\pm$ 0.26          &56.66 $\pm$ 0.25 \\ %
MTL~\cite{sun2019MTL}                     & \textcolor{green}{45.10 $\pm$ 1.8}          & 57.60 $\pm$ 0.9   \\  %
DC~\cite{lifchitz2019DC}                  & 42.04 $\pm$ 0.17         & 57.05 $\pm$ 0.16 \\  %
MetaOptNet~\cite{lee2019MetaOptNet}       & 41.10 $\pm$ 0.60         & 55.50 $\pm$ 0.60  \\ %
MatchNet~\cite{vinyals2016MatchNet}*      & 41.90 $\pm$ 0.27         & 54.41 $\pm$ 0.25 \\  %
TDE-FSL~\cite{Xing2022TDE}                & 44.61 $\pm$ 0.96         & 57.93 $\pm$ 0.81 \\  %
DeepEMD~\cite{zhang2020DeepEMD}*          & \textcolor{blue}{45.58 $\pm$ 0.26}      &\textcolor{red}{62.08 $\pm$ 0.25}  \\  %
DeepBDC~\cite{Xie2022DeepBDC}*            & 43.57 $\pm$ 0.25  &\textcolor{green}{59.49 $\pm$ 0.25}  \\ \hline  %
QSFormer (Ours)           & \textcolor{red}{46.51 $\pm$ 0.26}         &\textcolor{blue}{61.58 $\pm$ 0.25}              \\ \hline
\end{tabular}
\label{table:t2}
\end{table}

\begin{table}[!ht]
\centering
\caption{5-way result comparison of ours and state-of-the-art methods on Caltech-UCSD Birds-200-2011 dataset. The 1$^{st}$, 2$^{rd}$ and 3$^{rd}$ are respectively in \textcolor{red}{Red}, \textcolor{blue}{Blue} and \textcolor{green}{Green}.
* denotes this method is reproduced with our settings.}
\begin{tabular}{l|ll}
\hline
Method  & $\ \ \ \ \ $1-shot       & $\ \ \ \ \ $5-shot       \\ \hline
MELR~\cite{fei2021MELR}  & 70.26 $\pm$ 0.50   & 85.01 $\pm$ 0.32   \\ %
IEPT~\cite{zhang2021IEPT} & 69.97 $\pm$ 0.49   & 84.33 $\pm$ 0.33  \\  %
MVT~\cite{park2020MVT}    & \multicolumn{1}{c}{$\ \ -$}  & 85.35 $\pm$ 0.55   \\  %
FEAT~\cite{ye2019FEAT}*   & \textcolor{blue}{75.00 $\pm$ 0.29}   & \textcolor{blue}{86.24 $\pm$ 0.19}   \\ %
cosine classifier~\cite{chen2019cosine}* & 62.09 $\pm$ 0.29   & 80.04 $\pm$ 0.21 \\  %
ProtoNet~\cite{snell2017ProtoNet}* & \textcolor{green}{70.93 $\pm$ 0.30}   & 85.55 $\pm$ 0.19  \\ %
MatchNet~\cite{vinyals2016MatchNet}* & 70.21 $\pm$ 0.30   & 82.69 $\pm$ 0.22  \\  %
RelationNet~\cite{sung2018RelationNet} & 66.20 $\pm$ 0.99 & 82.30 $\pm$ 0.58 \\  %
MAML~\cite{finn2017MAML} & 67.28 $\pm$ 1.08   & 83.47 $\pm$ 0.59 \\  %
DEML~\cite{zhou2018DEML} & 66.95 $\pm$ 1.06   & 77.11 $\pm$ 0.78 \\  %
DeepEMD~\cite{zhang2020DeepEMD}* & 70.71 $\pm$ 0.30  &\textcolor{green}{86.13 $\pm$ 0.19}  \\  %
DeepBDC~\cite{Xie2022DeepBDC}* & 65.45 $\pm$ 0.29  &85.01 $\pm$ 0.19  \\ \hline %
QSFormer (Ours)               & \textcolor{red}{75.44 $\pm$ 0.29}   &\textcolor{red}{86.30 $\pm$ 0.19}         \\ \hline
\end{tabular}
\label{table:t3}
\end{table}

\begin{table}[!htbp]
\centering
\caption{Cross-domain experiments ($miniImagenet \to CUB$).
* denotes this method is reproduced with our settings.
The \textcolor{red}{red} represents the best results and \textcolor{blue}{blue} denotes the second-best results.}
\begin{tabular}{l|l|l}
\hline
Methods & 1-shot & 5-shot \\ \hline
ProtoNet~\cite{snell2017ProtoNet}    & 50.01 $\pm$ 0.82   & 72.02 $\pm$ 0.67     \\ 
MatchNet~\cite{vinyals2016MatchNet}        & 51.65 $\pm$ 0.84   & 69.14 $\pm$ 0.72     \\ 
cosine classifier~\cite{chen2019cosine}     & 44.17 $\pm$ 0.78   & 69.01 $\pm$ 0.74     \\ 
Baseline~\cite{chen2019cosine}              & \multicolumn{1}{c|}{$\ \ -$}   & 65.57 $\pm$ 0.70     \\ 
Baseline++~\cite{chen2019cosine}            & \multicolumn{1}{c|}{$\ \ -$}   & 62.04 $\pm$ 0.76     \\ 
FEAT~\cite{ye2019FEAT}*                    & 52.67 $\pm$ 0.29   & 72.65 $\pm$ 0.25     \\
DeepEMD~\cite{zhang2020DeepEMD}            &  \textcolor{blue}{54.24 $\pm$ 0.86}   & \textcolor{red}{78.86 $\pm$ 0.65}     \\
DeepBDC~\cite{Xie2022DeepBDC}*             & 50.28 $\pm$ 0.27   & 76.49 $\pm$ 0.23     \\ \hline
QSFormer (Ours)                    & \textcolor{red}{55.04 $\pm$ 0.29}     &  \textcolor{blue}{77.12 $\pm$ 0.24}                 \\ \hline
\end{tabular}
\label{table:t6}
\end{table}

\begin{table*}[!ht]
\centering
\small
\caption{Ablation study for the different components of the proposed QSFormer. The best results are highlighted in \textbf{bold}.
}
\setlength\tabcolsep{8pt}
\begin{tabular}{c|cccc|cccc}
\hline
&\multicolumn{4}{c|}{Different Components} &\multicolumn{4}{c}{Datasets}              \\ \cline{2-9}
\# &Baseline    &CIFE &sampleFormer &patchFormer                             & miniImageNet    &tieredImageNet   &FC100            &CUB    \\ \hline
1 &$\checkmark$ &  &  &                                    &59.64 $\pm$ 0.27 &55.87 $\pm$ 0.31 &39.47 $\pm$ 0.23 & 62.09 $\pm$ 0.29 \\
2 &$\checkmark$ &$\checkmark$ & &                          &61.15 $\pm$ 0.28 &70.73 $\pm$ 0.32 &41.54 $\pm$ 0.25 &65.95 $\pm$ 0.30 \\
3 &$\checkmark$ &$\checkmark$ &$\checkmark$ &              &63.97 $\pm$ 0.28 &71.64 $\pm$ 0.32 &45.46 $\pm$ 0.26 &72.93 $\pm$ 0.29 \\
4 &$\checkmark$ & $\checkmark$ &$\checkmark$ &$\checkmark$ &\textbf{65.24 $\pm$ 0.28} &\textbf{72.47 $\pm$ 0.31} &\textbf{46.51 $\pm$ 0.26} &\textbf{75.44 $\pm$ 0.29} \\ \hline
\end{tabular}
\label{table:t4}
\end{table*}

\begin{table*}[!ht]
\centering
\caption{ Hyperparameter settings of our proposed QSFormer.}
\begin{tabular}{c|ccccc}
\hline
\multirow{2}{*}{Hyper-parameters} & \multicolumn{5}{c}{Datasets}    \\ 
                             &miniImageNet       &tieredImageNet     &FC100            &CUB              &miniImageNet $\to$ CUB \\ \hline
Optimizer                    & SGD               & SGD               & SGD             & SGD             & SGD         \\ 
Initial LR                   & 5e-4              & 5e-4              & 1e-4            & 5e-4            & 5e-4         \\ 
Steps of LR decay            & 10                & 10                & 10              & 10              & 10           \\ 
Coefficient of LR decay      & 0.9               & 0.5               & 0.9             & 0.95            & 0.9          \\ 
N                            & 3                 & 3                 & 4               & 2               & 3            \\ 
Number of Head               & 10,8              & 8,8               & 8,1             & 8,1             & 10,8         \\ 
dropout rates                & 0.5,0.5,0.5,0.1   & 0.5,0.5,0.5,0.1   & 0.5,0.5,0.5,0.1 & 0.1,0.5,0.5,0.1 & 0.5,0.5,0.5,0.1   \\ 
$\alpha$                     & 0.7               & 0.5               & 0.5             & 0.05            & 0.7           \\ 
$\lambda$                    & 0.1               & 0.1               & 0.4             & 0.3             & 0.1            \\ 
Epochs                       & 100               & 100               & 50              & 150             & 100           \\ \hline
\end{tabular}
\label{table:t7}
\end{table*}

\begin{table*}[!htbp]
\centering
\caption{Performance comparison of the classical methods based on different metric learning.
* denotes the comparison methods is reproduced with our setting.
The \textbf{bold black} represents the best results.}
\begin{tabular}{c|c|cccc}
\hline
Methods        &Metric       &miniImageNet      &tieredImageNet    &FC100            &CUB \\ \hline
cosine classifier~\cite{chen2019cosine}* &Cosine   &59.64 $\pm$ 0.27  &55.87 $\pm$ 0.31  &39.47 $\pm$ 0.23 &62.09 $\pm$ 0.29  \\
MatchNet~\cite{vinyals2016MatchNet}*     &Cosine   &61.24 $\pm$ 0.29  &71.01 $\pm$ 0.33  &41.90 $\pm$ 0.27 &70.20 $\pm$ 0.30  \\
ProtoNet~\cite{snell2017ProtoNet}*      &Euclidean       &63.03 $\pm$ 0.29  &68.68 $\pm$ 0.34  &40.91 $\pm$ 0.26 &70.93 $\pm$ 0.30  \\
DeepEMD~\cite{zhang2020DeepEMD}*   &EMD  &\textbf{65.43 $\pm$ 0.28}  &69.84 $\pm$ 0.32  &45.58 $\pm$ 0.26 &70.71 $\pm$ 0.30  \\ \hline
QSFormer       &Ours   &65.24 $\pm$ 0.28  &\textbf{72.47 $\pm$ 0.31}  &\textbf{46.51 $\pm$ 0.26} &\textbf{75.44 $\pm$ 0.29}   \\ \hline
\end{tabular}
\label{table:t8}
\end{table*}

\subsection{Ablation Study}
To better understand the effectiveness of our proposed QSFormer, in this section, we conduct extensive ablation studies, including component analysis, similarity metric analysis, cross-domain analysis, etc.

\textbf{Component Analysis.}
Our proposed QSFormer mainly contains three components: Cross-scale Interactive Feature Extractor (CIFE), Sample Transformer Module (sampleFormer) and Patch Transformer Module (patchFormer).
The experimental results of ablation study are shown in Table~\ref{table:t4}.
We reproduce cosine classifier method~\cite{chen2019cosine} consisting of CNN network and cosine distance as the Baseline network for comparison.
From Table~\ref{table:t4}, we can observe:
(1) By comparing \#1 with \#2, the performance of Baseline network can be significantly improved with the help of CIFE, which demonstrates the effectiveness of CIFE.
(2) By comparing \#2 with \#3, we can find that sampleFormer significantly improves the performance of model based on \#2, which indicates the effectiveness of sampleFormer module.
(3) By adding patchFormer into \#3, we further improve the performance of whole network, which shows the effectiveness of patchFormer module.
All these experiments fully validate the effectiveness of each component in our proposed QSFormer framework.

\begin{figure}[!ht]
\centering
\subfigure[Baseline]{\includegraphics[width=4.1cm]{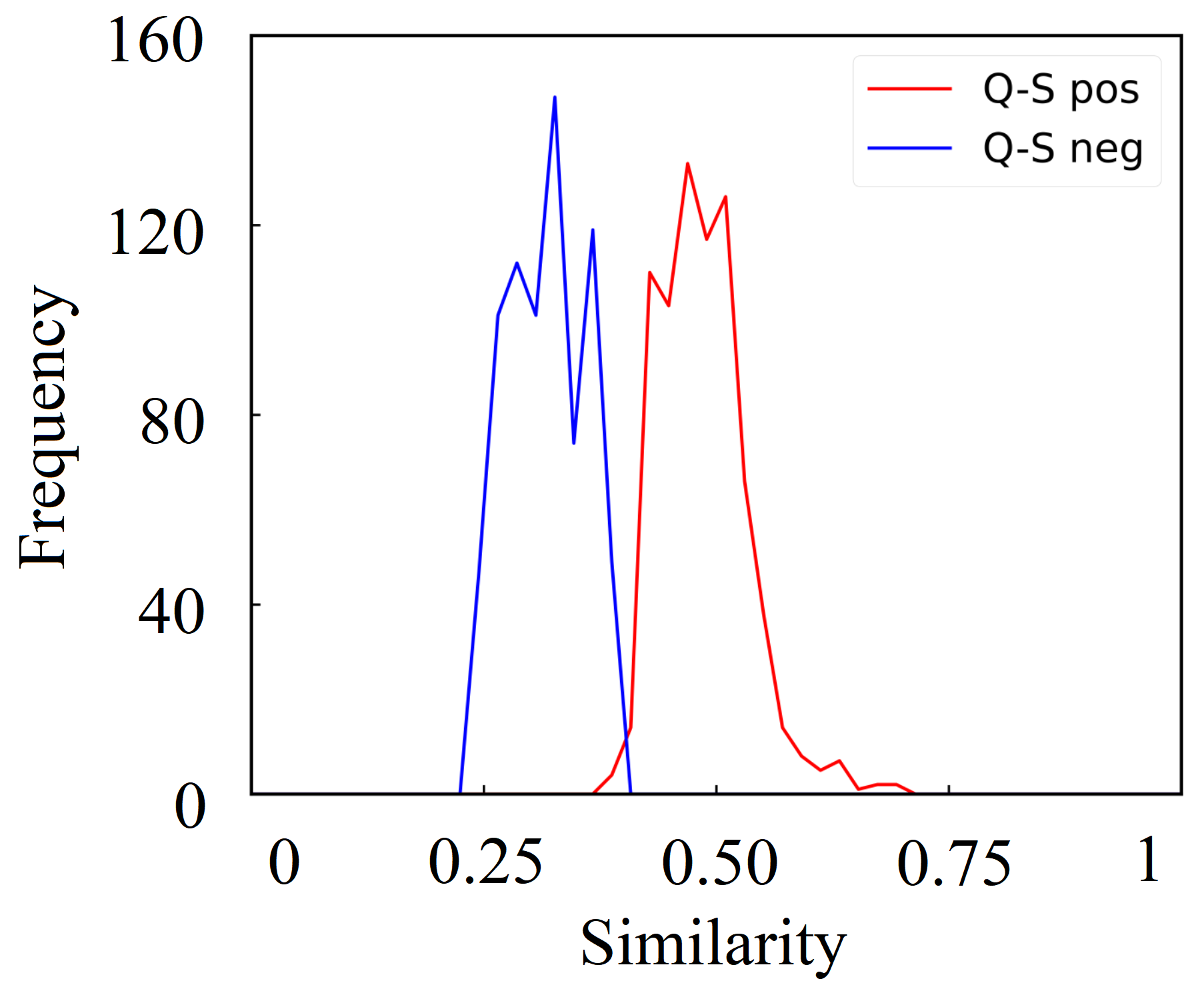}}
\subfigure[QSFormer]{\includegraphics[width=4.1cm]{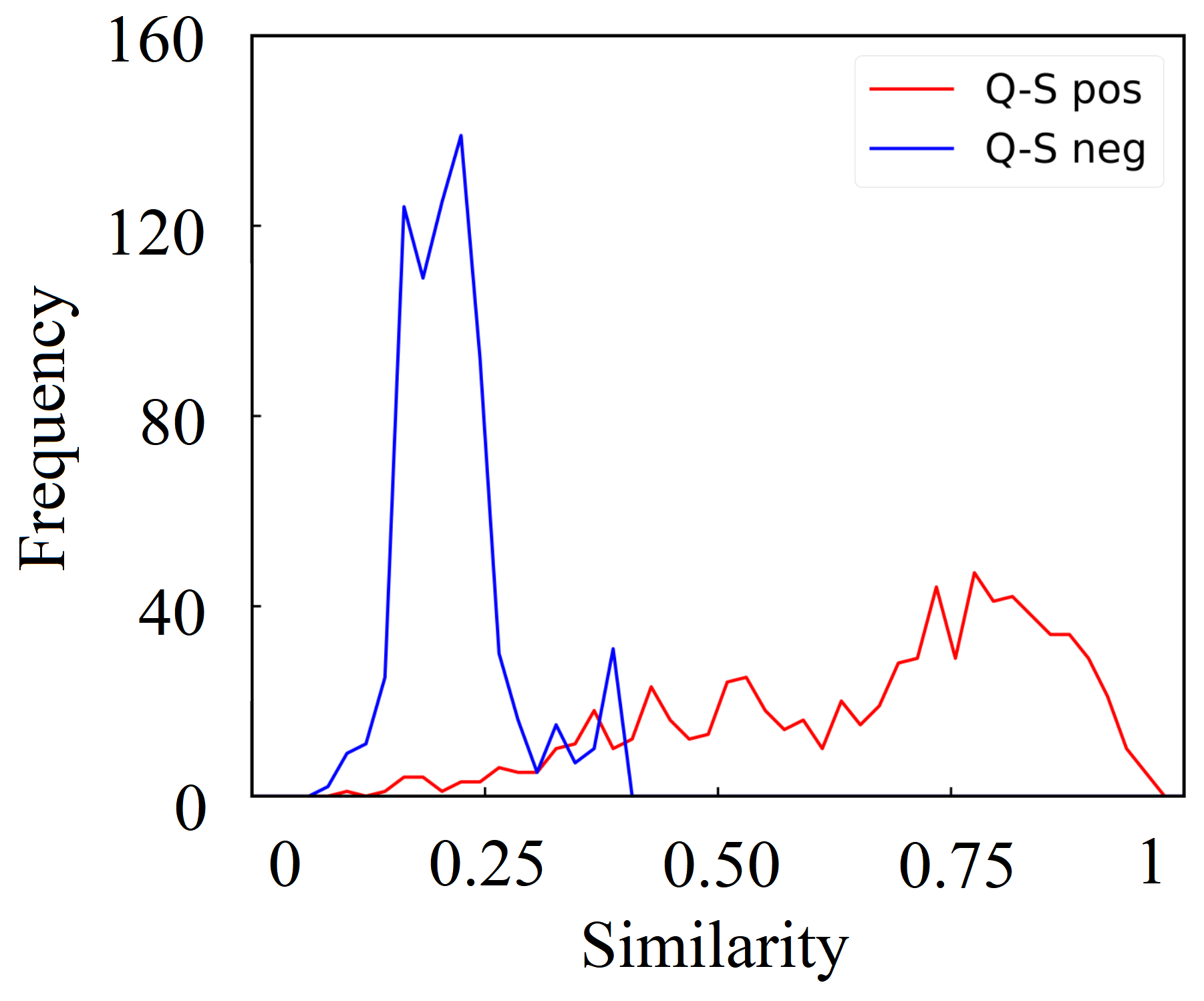}}
\caption{Comparison of similarity distribution between Baseline and our QSFormer.
The similarities of ``Q-S pos" become larger while the similarities of ``Q-S neg" become smaller, which indicates they are more easily separated.
} 
\label{fig:fig5}
\end{figure}

\textbf{Similarity Metric Analysis.}
To verify the effectiveness of the proposed QSFormer on metric learning, we visualize the similarity distribution of Baseline and QSFormer on the more challenging 5-way 1-shot task, as shown in Figure~\ref{fig:fig5}.
For 5-way 1-shot task, 
each query sample generates the similarity results of one positive query-support sample pair (i.e., ``Q-S pos") and four negative query-support sample pairs (i.e., ``Q-S neg") during the metric learning process.
To facilitate the comparison of the similarity results of ``Q-S pos" and ``Q-S neg", we average the similarity values of four ``Q-S neg" corresponding to each query sample.
For this experiment, we perform 10 episodes, where each episode random selects $15 \times 5 = 75$ query samples for classification, i.e., we can get the $75 \times 10 = 750$ similarity values of ``Q-S pos" and ``Q-S neg", respectively.
Subsequently,
we count the number of ``Q-S pos" and ``Q-S neg" within a certain range according to the normalized similarity values and thus produce the similarity distribution as shown in Figure~\ref{fig:fig5}.
We can observe that:
(1) the similarity values of ``Q-S pos" obtained by the Baseline method are generally below 0.5, while ``Q-S neg" are above 0.25.
(2) In our proposed QSFormer, the similarity values of ``Q-S pos" are mostly above 0.5, while ``Q-S neg" are mostly below 0.25.
Therefore, our proposed QSFormer can separate positive and negative query-support sample pairs more accurately.

In addition, we also compare our QSFormer with other metric learning algorithms, including cosine classifier~\cite{chen2019cosine}, MatchNet~\cite{vinyals2016MatchNet}, ProtoNet~\cite{snell2017ProtoNet} and DeepEMD~\cite{zhang2020DeepEMD}.
These compared methods are reproduced with the same settings and training schemes as ours for a more fair comparison.
As shown in Table~\ref{table:t8}, we can observe that our proposed method obtains the best performance on four publicly popular datasets, which fully demonstrates the effectiveness and superiority of our proposed QSFormer.
These experiments fully demonstrate the effectiveness of our proposed QSFormer for metric learning.


\textbf{Cross-domain Analysis. }
To validate the transferable ability of our proposed QSFormer, we conduct a cross-domain experiment by following~\cite{chen2019cosine, zhang2020DeepEMD}. The training and testing are implemented on miniImagenet dataset and CUB dataset, respectively. As shown in Table~\ref{table:t6}, our proposed QSFormer achieves the best performance on the 1-shot setting (55.04 $\pm$ 0.29) and the second-best results on the 5-shot, i.e., 77.12 $\pm$ 0.24.
%
These results demonstrate that the proposed QSFormer learns the discriminative information across domains, and adaptively explores the correspondence of query-support samples.

\begin{figure}[!t]
\centering
\includegraphics[width=0.48\textwidth]{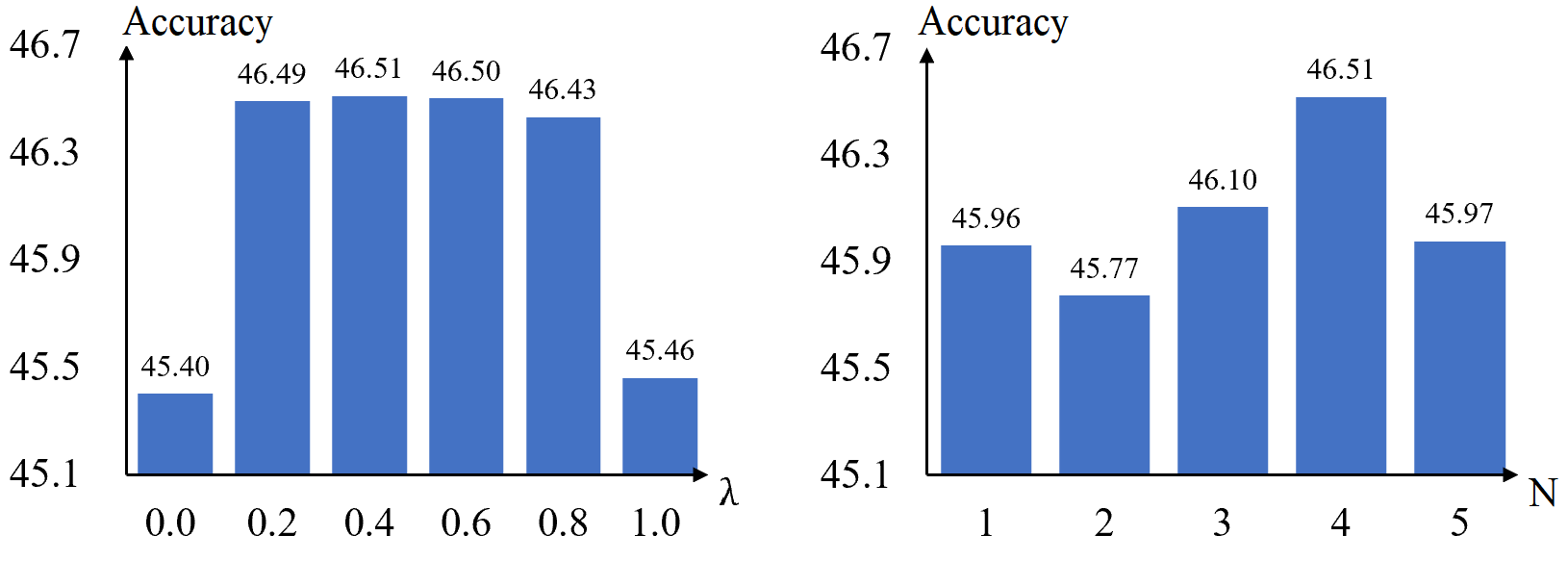}
\caption{Ablation study of two parameters (i.e., $\lambda$ and $N$).
}
\label{fig:fig4}
\end{figure}

\textbf{Parameter Analysis.}
There are two important parameters in our model, including the balanced parameter $\lambda$ in Equ. (\ref{equ:eq12}) for local and global metric, and the number of sampleFormer layers $N$. In this section, we conduct experiments on the FC100 dataset on 5-way 1-shot task to check their influence. As shown in Figure~\ref{fig:fig4}, we can observe that the performance is relatively stable when we slightly adjust the balanced parameter $\lambda$ in the range of (0.2, 0.6). For the number $N$ of sampleFormer layers, we can find that our performance is increasing continuously when the $N$ is changing from 2 to 4. Therefore, we set $\lambda=0.4$ and $N=4$ for our experiments.


\section{Conclusion}
In this paper, we propose a novel unified Query-Support Transformer (QSFormer)  to deeply exploit the  sample relationships in query and support sets for few-shot classification task. QSFormer mainly contains sample  Transformer (sampleFormer)  module and patch Transformer  (patchFormer) module.
sampleFormer is designed to meet the problem setting of few-shot classification, i.e., it couples the sample representation and metric learning between query and support sets together via a single
Transformer architecture.
Meanwhile, as a complementary,  patchFormer is also adopted to model the local structural metric between query and support samples.
A new CNN feature extractor (CIFE) is also proposed to provide an effective CNN backbone for our approach.
Extensive experiments demonstrate the effectiveness and superiority of our proposed QSFormer approach.

\bibliographystyle{IEEEtran}
\bibliography{reference}

\end{document}